\documentclass[11pt,a4paper]{article}
\usepackage[hyperref]{acl2020}
\usepackage{times}
\usepackage{latexsym}

\usepackage{multirow}
\usepackage{booktabs}
\usepackage{xcolor}
\usepackage{amssymb}
\usepackage{graphicx}
\usepackage{amsmath}
\usepackage{enumitem}

\usepackage{pifont}% http://ctan.org/pkg/pifont
\newcommand{\cmark}{\ding{51}}%
\newcommand{\xmark}{\ding{55}}%
\usepackage{microtype}

\aclfinalcopy % Uncomment this line for the final submission
\title{TVQA+: Spatio-Temporal Grounding for Video Question Answering}

\author{
  Jie Lei $\;\;\;\;\;$ Licheng Yu $\;\;\;\;\;$ 
  Tamara L. Berg $\;\;\;\;\;$ Mohit Bansal \\
  Department of Computer Science \\ University of North Carolina at Chapel Hill \\
  {\tt \{jielei, licheng, tlberg, mbansal\}@cs.unc.edu} \\
}

\begin{document}
\maketitle
\begin{abstract}
We present the task of Spatio-Temporal Video Question Answering, which requires intelligent systems to simultaneously retrieve relevant moments and detect referenced visual concepts (people and objects) to answer natural language questions about videos. 
We first augment the TVQA dataset 
with 310.8K bounding boxes, linking depicted objects to visual concepts in questions and answers.
We name this augmented version as TVQA+.
We then propose Spatio-Temporal Answerer with Grounded Evidence (STAGE), a unified framework that grounds evidence in both spatial and temporal domains to answer questions about videos.
Comprehensive experiments and analyses demonstrate the effectiveness of our framework and how the rich annotations in our TVQA+ dataset can contribute to the question answering task.
Moreover, by performing this joint task, our model is able to produce insightful and interpretable spatio-temporal attention visualizations.\footnote{Dataset and code are publicly available: \url{http://tvqa.cs.unc.edu}, \url{https://github.com/jayleicn/TVQAplus}}
\end{abstract}

\section{Introduction}\label{intro}

We have witnessed great progress in recent years on image-based visual question answering (QA) tasks~\cite{Antol2015VQAVQ,Yu2015VisualMF,Zhu2016Visual7WGQ}.
One key to this success has been spatial attention~\cite{Anderson2017BottomUpAT,Shih2016WhereTL,Lu2016HierarchicalQC}, where neural models learn to attend to relevant regions for predicting the correct answer.
Compared to image-based QA, there has been less progress on the performance of video-based QA tasks.
One possible reason is that attention techniques are hard to generalize to the temporal nature of videos.
Moreover, due to the high cost of annotation, most existing video QA datasets only contain QA pairs, without providing labels for the key clips or regions needed to answer the question.
Inspired by previous work on grounded image and video captioning~\cite{Lu2018NeuralBT,Zhou2018GroundedVD}, we propose methods that explicitly localize video clips as well as spatial regions for answering video-based questions. 
Such methods are useful in many scenarios, such as natural language guided spatio-temporal localization, and adding explainability to video question answering, which is potentially useful for decision making and model debugging.
To enable this line of research, we also collect new joint spatio-temporal annotations for an existing video QA dataset.

\begin{figure}[!t]
  \centering
  \includegraphics[width=0.95\linewidth]{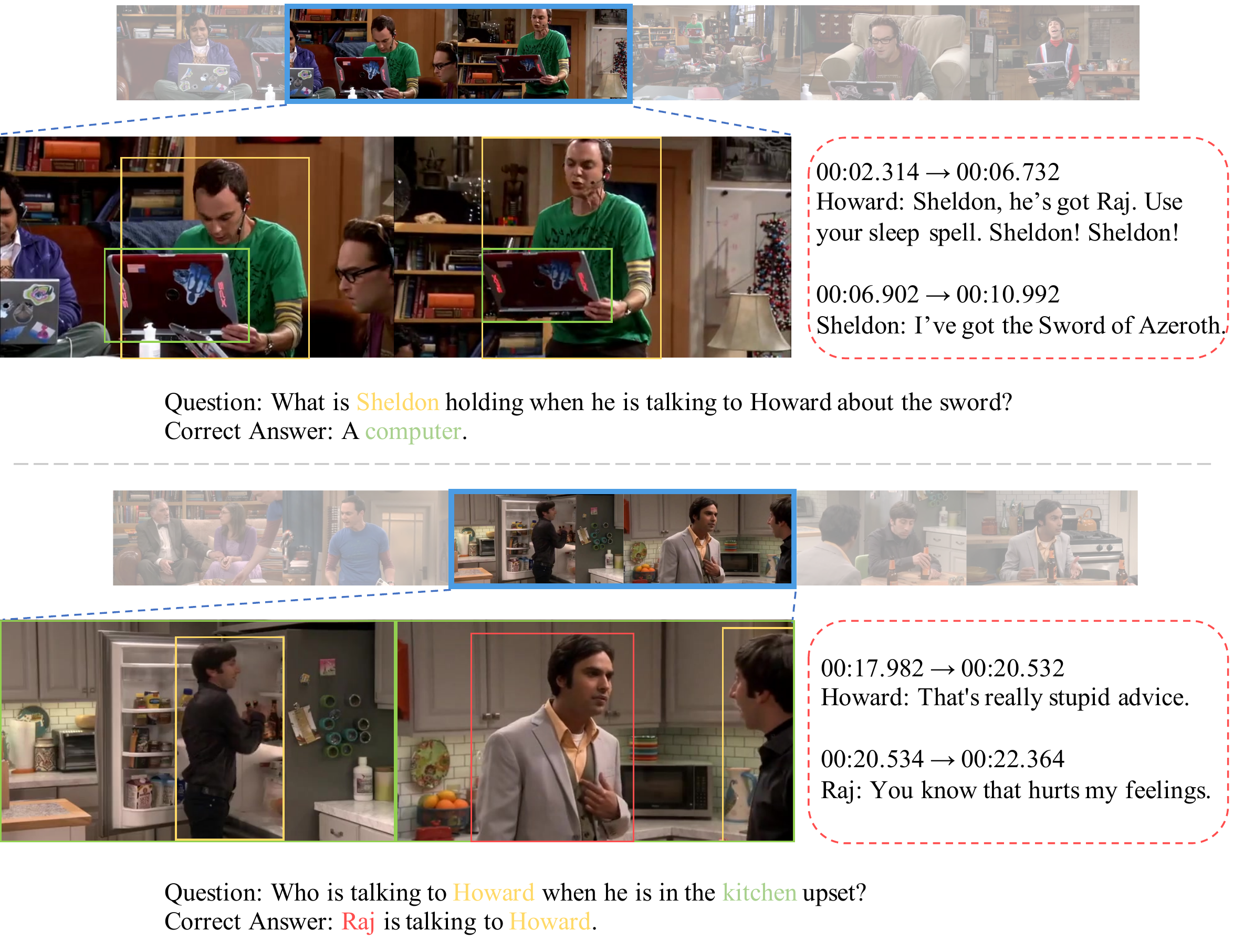}
  \caption{Samples from TVQA+. Questions and correct answers are temporally localized to clips, and spatially localized to frame-level bounding boxes. Colors indicate corresponding box-object pairs. Subtitles are shown in dashed blocks. Wrong answers are omitted.}
  \label{fig:qa_box_example}
\end{figure}

In the past few years, several video QA datasets have been proposed, e.g., MovieFIB~\cite{Maharaj2017ADA}, MovieQA~\cite{Tapaswi2016MovieQAUS}, TGIF-QA~\cite{Jang2017TGIFQATS}, PororoQA~\cite{Kim2017DeepStoryVS}, MarioQA~\cite{mun2017marioQA}, and TVQA~\cite{lei2018tvqa}.
TVQA is one of the largest video QA datasets, providing a large video QA dataset built on top of 6 famous TV series. 
Because TVQA was collected on television shows, it is built on natural video content with rich dynamics and complex social interactions, where question-answer pairs are written by people observing both videos and their accompanying dialogues, encouraging the questions to require both vision and language understanding to answer. 
Movie~\cite{Tapaswi2016MovieQAUS,Maharaj2017ADA} and television show~\cite{lei2018tvqa} videos come with the limitation of being scripted and edited, but they are still more realistic than cartoon/animation~\cite{Kim2017DeepStoryVS} and game~\cite{mun2017marioQA} videos, and they also come with richer, real-world-inspired inter-human interactions and span across diverse domains (e.g., medical, crime, sitcom, etc.), making them a useful testbed to study complex video understanding by machine learning models.

One key property of TVQA is that it provides temporal annotations denoting which parts of a video clip are necessary for answering a proposed question.
However, none of the existing video QA datasets (including TVQA) provide spatial annotation for the answers.
Actually, grounding spatial regions correctly could be as important as grounding temporal moments for answering a given question.
For example, in Fig.~\ref{fig:qa_box_example}, to answer the question of \textit{``What is Sheldon holding when he is talking to Howard about the sword?''}, we need to localize the moment when \textit{``he is talking to Howard about the sword?''}, as well as look at the region of \textit{``What is Sheldon holding''}.

Hence, in this paper, we first augment a subset of the TVQA dataset with grounded bounding boxes, resulting in a spatio-temporally grounded video QA dataset, TVQA+.
It consists of 29.4K multiple-choice questions grounded in both the temporal and the spatial domains.
To collect spatial groundings, we start by identifying a set of visual concept words, i.e., objects and people, mentioned in the question or correct answer. 
Next, we associate the referenced concepts with object regions in individual frames, if there are any, by annotating bounding boxes for each referred concept (see examples in Fig.~\ref{fig:qa_box_example}). 
Our TVQA+ dataset has a total of 310.8K bounding boxes linked with referred objects and people, spanning across 2.5K categories (more details in Sec.~\ref{dataset}).

With such richly annotated data, we then propose the task of spatio-temporal video question answering, which requires intelligent systems to localize relevant moments, detect referred objects and people, and answer questions. 
We further design several metrics to evaluate the performance of the proposed task, including QA accuracy, object grounding precision, temporal localization accuracy, and a joint temporal localization and QA accuracy.
To address spatio-temporal video question answering, we propose a novel end-to-end trainable model, Spatio-Temporal Answerer with Grounded Evidence (STAGE), 
which effectively combines moment localization, object grounding, and question answering in a unified framework.
We find that the QA performance benefits from both temporal moment and spatial region supervision. 
Additionally, we provide visualization of temporal and spatial localization, which is helpful for understanding what our model has learned.
Comprehensive ablation studies demonstrate how each of our annotations and model components helps to improve the performance of the tasks. 

\smallskip
\noindent To summarize, our contributions are:
\begin{itemize}[leftmargin=*]
    \item We collect TVQA+, 
    a large-scale spatio-temporal video question answering dataset, which augments the original TVQA dataset with frame-level bounding box annotations. To our knowledge, this is the first dataset that combines moment localization, object grounding, and question answering. 
    \item We design a novel video question answering framework, Spatio-Temporal Answerer with Grounded Evidence (STAGE),  to jointly localize moments, ground objects, and answer questions. By performing all three sub-tasks together, our model achieves significant performance gains over the baselines, as well as presents insightful, interpretable visualizations.
\end{itemize}

%%%%%%%%%%%%%%%%%%%%%%%%%%%%%%%%%%%%%%
% Section 2 Related Work
%%%%%%%%%%%%%%%%%%%%%%%%%%%%%%%%%%%%%%

%%%%%%%%%%%%%%%%%%%%%%%%%%%%%%%%%%%%%%
% Section 3 Dataset
%%%%%%%%%%%%%%%%%%%%%%%%%%%%%%%%%%%%%%

\begin{table*}[!ht]
\centering
\small
\scalebox{0.99}{
\begin{tabular}{lllrcc}
\toprule
\multirow{2}{*}{Dataset} & \multirow{2}{*}{ Origin} & \multirow{2}{*}{Task} & \#Clips/\#QAs  & \multirow{2}{*}{\#Boxes} & Temporal  \\ 
 & & & (\#Sentences) & & Annotation  \\ 
\midrule
MovieFIB~\cite{Maharaj2017ADA} & Movie & QA & 118.5K/349K & - & \xmark \\
MovieQA~\cite{Tapaswi2016MovieQAUS} & Movie & QA & 6.8K/6.5K & - & \cmark \\
TGIF-QA~\cite{Jang2017TGIFQATS} & Tumblr & QA & 71.7K/165.2K & - & \xmark \\
PororoQA~\cite{Kim2017DeepStoryVS} & Cartoon & QA & 16.1K/8.9K & - & \xmark \\
DiDeMo~\cite{Hendricks2017LocalizingMI} & Flickr & TL & 10.5K/40.5K & - & \cmark \\
Charades-STA~\cite{Gao2017TALLTA} & Home & TL & -/19.5K & - & \cmark \\ 
TVQA~\cite{lei2018tvqa} & TV Show \ \ \ & QA/TL & 21.8K/152.5K & - & \cmark \\ 
ANet-Entities~\cite{Zhou2018GroundedVD} \ \ \ \ & Youtube & CAP/TL/SL & 15K/52K & 158K & \cmark \\ 
\midrule
TVQA+ & TV Show & QA/TL/SL & 4.2K/29.4K & 310.8K & \cmark \\ 
\bottomrule
\\
\end{tabular}
}
\vspace{-10pt}
\caption{Comparison of TVQA+ with other video-language datasets. TL=Temporal Localization, SL=Spatial Localization, CAP=Captioning.
}
\label{tab:dataset_comparison}
\end{table*}

\section{Related Work}\label{related}

\paragraph{Question Answering}  
In recent years, multiple question answering datasets and tasks have been proposed to facilitate research towards this goal, in both vision and language communities, in the form of visual question answering~\cite{Antol2015VQAVQ,Yu2015VisualMF,Jang2017TGIFQATS} and textual question answering~\cite{Rajpurkar2016SQuAD10,Weston2016TowardsAQ}, respectively. 
Video question answering~\cite{lei2018tvqa,Tapaswi2016MovieQAUS,Kim2017DeepStoryVS} with naturally occurring subtitles are particularly interesting, as it combines both visual and textual information for question answering. 
Different from existing video QA tasks, where a system is only required to predict an answer, we propose a novel task that additionally grounds the answer in both spatial and temporal domains.

\paragraph{Language-Guided Retrieval} 
Grounding language in images/videos is an interesting problem that requires jointly understanding both text and visual modalities. Earlier works~\cite{Kazemzadeh2014ReferItGameRT,yu2017joint,Yu_2018_CVPR,Rohrbach2016GroundingOT} focused on identifying the referred object in an image. 
Recently, there has been a growing interest in moment retrieval tasks~\cite{Hendricks2017LocalizingMI,Hendricks2018LocalizingMI,Gao2017TALLTA}, where the goal is to localize a short clip from a long video via a natural language query. 
Our work integrates the goals of both tasks, requiring a system to ground the referred moments and objects simultaneously.

\paragraph{Temporal and Spatial Attention}
Attention has shown great success on many vision and language tasks, such as image captioning~\cite{Anderson2017BottomUpAT,Xu2015ShowAA}, visual question answering~\cite{Anderson2017BottomUpAT,Trott2018InterpretableCF}, language grounding~\cite{Yu_2018_CVPR}, etc. 
However, sometimes the attention learned by the model itself may not agree with human expectations~\cite{Liu2016AttentionCI,Das2016HumanAI}. 
Recent works on grounded image captioning and video captioning~\cite{Lu2018NeuralBT,Zhou2018GroundedVD} show better performance can be achieved by explicitly supervising the attention. 
In this work, we use annotated frame-wise bounding box annotations to supervise both temporal and spatial attention.
Experimental results demonstrate the effectiveness of supervising both domains in video QA.

\begin{table}[t]
\centering
\small
\scalebox{0.86}{
\begin{tabular}{lrrrrr}
\toprule
\multirow{2}{*}{Split}    & \multirow{2}{*}{\#Clips/\#QAs}  & \#Annotated & \multirow{2}{*}{\#Boxes} & \multirow{2}{*}{\#Categories}  \\ 
    &   & Images &  &   \\ 
\midrule
Train & 3,364/23,545   & 118,930 & 249,236 & 2,281 \\
Val & 431/3,017  & 15,350 & 32,682 & 769 \\
Test & 403/2,821 & 14,188 & 28,908 & 680 \\ 
\midrule
Total & 4,198/29,383 & 148,468 & 310,826 & 2,527  \\
\bottomrule
\end{tabular}
}
\caption{Data Statistics for TVQA+ dataset.}
\label{tab:data_stat_main}
\vspace{-5pt}
\end{table}

\begin{figure*}[t]
\centering
  \includegraphics[width=0.92\linewidth]{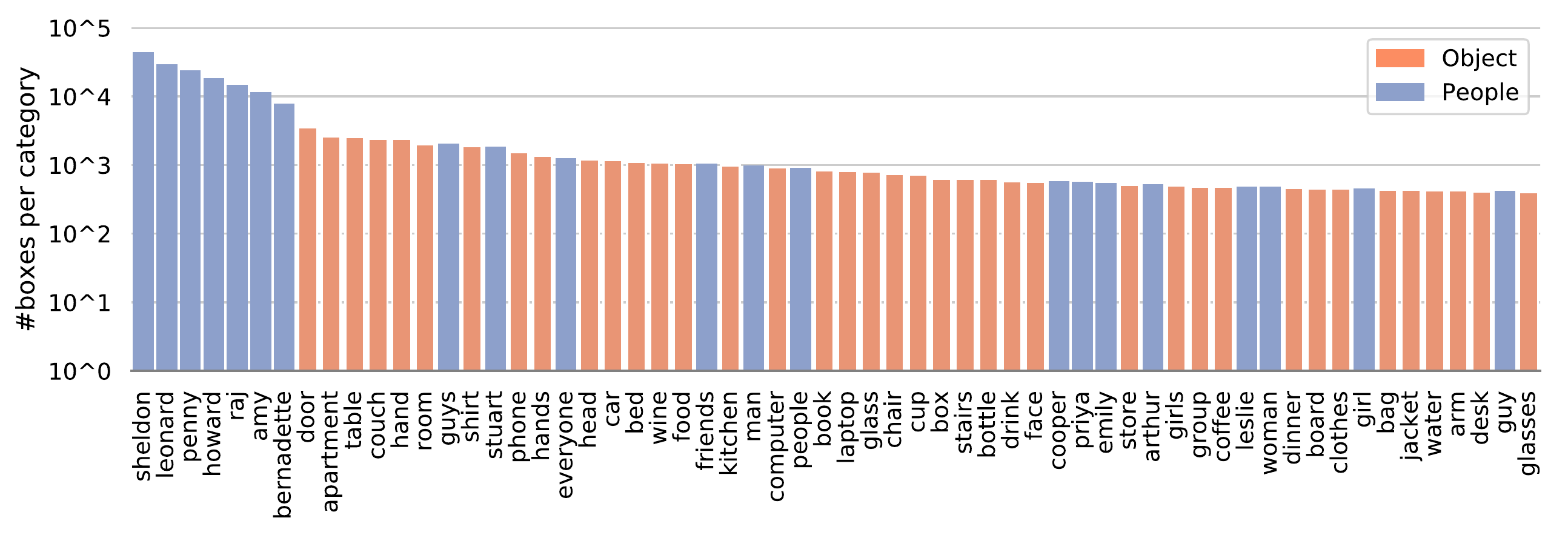}
  \caption{Box distributions for top 60 categories in TVQA+ train set.}
  \label{fig:cat_dist}
\end{figure*}

\section{Dataset}\label{dataset}
In this section, we describe the TVQA+ Dataset, the first video question answering dataset with both spatial and temporal annotations. 
TVQA+ is built on the TVQA dataset introduced by~\citeauthor{lei2018tvqa}. 
TVQA is a large-scale video QA dataset based on 6 popular TV shows, containing 152.5K multiple choice questions from 21.8K, 60-90 second long video clips. 
The questions in the TVQA dataset are compositional, where each question is comprised of two parts, a question part (``where was Sheldon sitting''), joined via a link word, (``before'', ``when'', ``after''), to a localization part that temporally locates when the question occurs (``he spilled the milk'').
Models should answer questions using both visual information from the video, as well as language information from the naturally associated dialog (subtitles).
Since the video clips on which the questions were collected are usually much longer than the context needed for answering the questions, the TVQA dataset also provides a temporal timestamp annotation indicating the minimum span (context) needed to answer each question. 
While the TVQA dataset provides a novel question format and temporal annotations, it lacks spatial grounding information, i.e., bounding boxes of the concepts (objects and people) mentioned in the QA pair. 
We hypothesize that object annotations could provide an additional useful training signal for models to learn a deeper understanding of visual information. 
Therefore, to complement the original TVQA dataset, we collect frame-wise bounding boxes for visual concepts mentioned in the questions and correct answers. 
Since the full TVQA dataset is very large, we start by collecting bounding box annotations for QA pairs associated with \textit{The Big Bang Theory}. 
This subset contains 29,383 QA pairs from 4,198 clips.

\subsection{Data Collection}
\paragraph{Identify Visual Concepts} 
To annotate the visual concepts in video frames, the first step is to identify them in the QA pairs. 
We use the Stanford CoreNLP part-of-speech tagger~\cite{manning-EtAl:2014:P14-5} to extract all nouns in the questions and correct answers. 
This gives us a total of 152,722 words from a vocabulary of 9,690 words. 
We manually label the non-visual nouns (e.g., ``plan'', ``time'', etc.) in the top 600 nouns, removing 165 frequent non-visual nouns from the vocabulary.

\paragraph{Bounding Box Annotation} 
For the selected \textit{The Big Bang Theory} videos from TVQA, we first ask Amazon Mechanical Turk workers to adjust the start and end timestamps to refine the temporal annotation, as we found the original temporal annotation were not ideally tight.
We then sample one frame every two seconds from each span for spatial annotation.
For each frame, we collect the bounding boxes for the visual concepts in each QA pair.  
We also experimented with semi-automated annotation for people with face detection~\cite{zhang2016joint} and recognition model~\cite{liu2017sphereface}, but they do not work well mainly due to many partial occlusion of faces (e.g., side faces) in the frames.
During annotation, we provide the original videos (with subtitles) to help the workers understand the context for the given QA pair. 
More annotation details (including quality check) are presented in the appendix.

\begin{figure}[t]
  \includegraphics[width=0.98\linewidth]{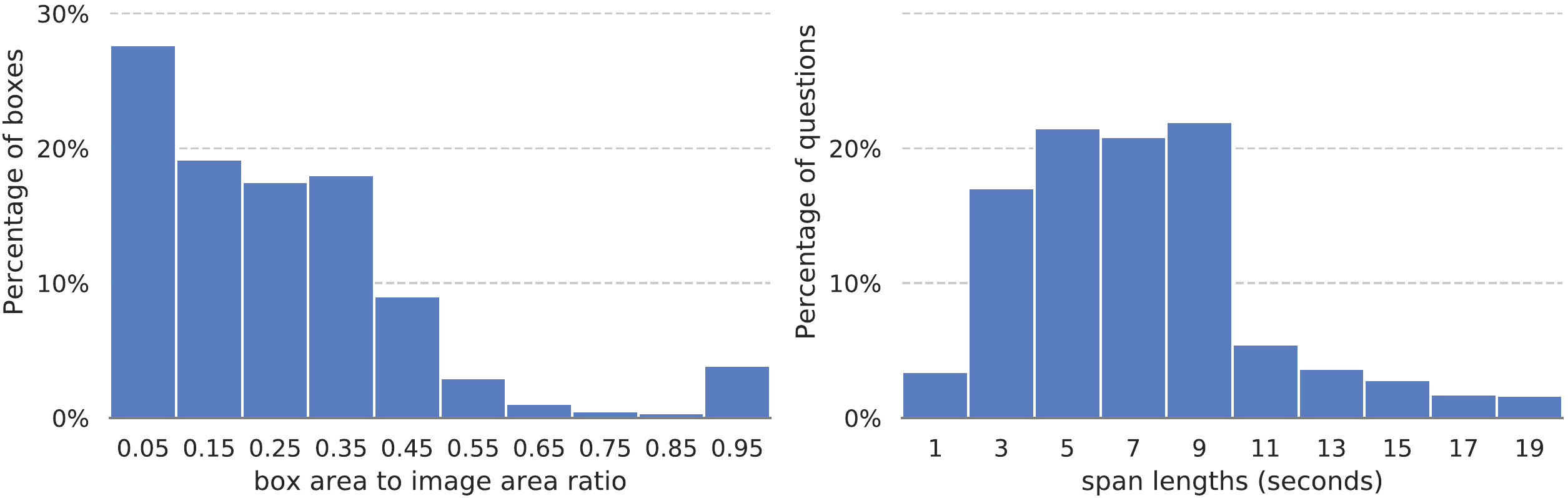}
  \caption{Box/image area ratios (left) and span length distributions (right) in TVQA+.}
  \label{fig:box_area_span_len}
\end{figure}

\subsection{Dataset Analysis}
TVQA+ contains 29,383 QA pairs from 4,198 videos, with 148,468 images annotated with 310,826 bounding boxes. 
Statistics of TVQA+ are shown in Table~\ref{tab:data_stat_main}. 
Note that we follow the same data splits as the original TVQA dataset, supporting future research on both TVQA and TVQA+.
Table~\ref{tab:dataset_comparison} compares TVQA+ dataset with other video-language datasets. 
TVQA+ is unique as it supports three tasks: question answering, temporal localization, and spatial localization. 

\begin{figure*}[!ht]
\centering
  \includegraphics[width=0.96\linewidth]{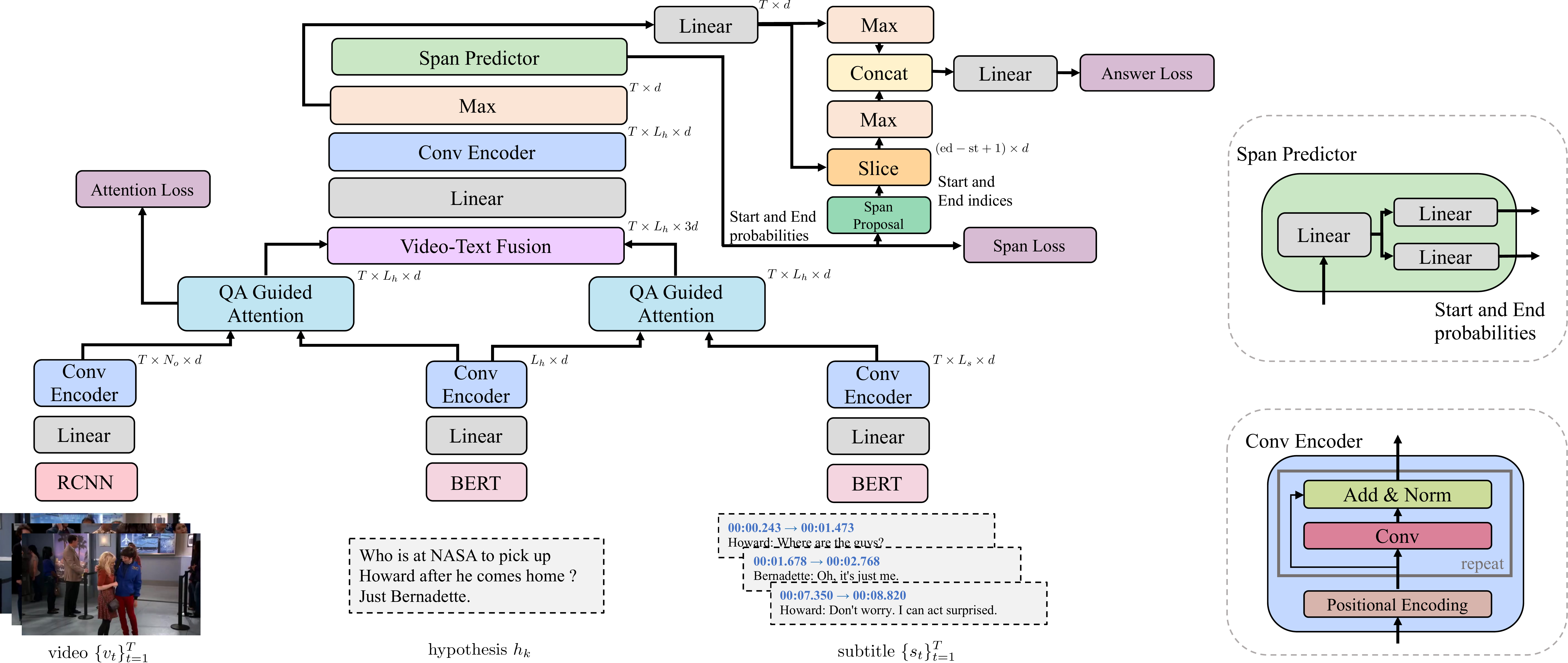}
  \caption{Overview of the proposed STAGE framework.}
  \label{fig:model_main}
\end{figure*}

It is also of reasonable size compared to the grounded video captioning dataset ANet-Entities~\cite{Zhou2018GroundedVD}.
On average, we obtain 2.09 boxes per image and 10.58 boxes per question. 
The annotated boxes cover 2,527 categories. 
We show the number of boxes (in log scale) for each of the top 60 categories in Fig.~\ref{fig:cat_dist}. 
The distribution has a long tail, e.g., the number of boxes for the most frequent category ``sheldon'' is around 2 orders of magnitude larger than the 60th category ``glasses''. 
We also show the distribution of bounding box area over image area ratio in Fig.~\ref{fig:box_area_span_len} (left). 
The majority of boxes are fairly small compared to the image, which makes object grounding challenging. 
Fig.~\ref{fig:box_area_span_len} (right) shows the distribution of localized span length. 
While most spans are less than 10 seconds, the largest spans are up to 20 seconds.
The average span length is 7.2 seconds, which is short compared to the average length of the full video clips (61.49 seconds).

%%%%%%%%%%%%%%%%%%%%%%%%%%%%%%%%%%%%%%
% Section 4 Methods
%%%%%%%%%%%%%%%%%%%%%%%%%%%%%%%%%%%%%%
\section{Methods}\label{sec:methods}

Our proposed method, Spatio-Temporal Answerer with Grounded Evidence (STAGE), is a unified framework for moment localization, object grounding and video QA.
First, STAGE encodes the video and text (subtitle, QA) via frame-wise regional visual representations and neural language representations, respectively. 
The encoded video and text representations are then contextualized using a Convolutional Encoder.
Second, STAGE computes attention scores from each QA word to object regions and subtitle words. 
Leveraging the attention scores, STAGE is able to generate QA-aware representations, as well as automatically detecting the referred objects/people. 
The attended QA-aware video and subtitle representation are then fused together to obtain a joint frame-wise representation.
Third, taking the frame-wise representation as input, STAGE learns to predict QA relevant temporal spans, then combines the global and local (span localized) video information to answer the questions. 
In the following, we describe STAGE in detail.

\subsection{Formulation}\label{sec:method_formulation}
In our tasks, the inputs are: (1) a question with 5 candidate answers; (2) a 60-second long video; (3) a set of subtitle sentences. 
Our goal is to predict the answer and ground it both spatially and temporally.
Given the question, $q$, and the answers, $\{a_{k}\}_{k=1}^{5}$, we first formulate them as 5 hypotheses (QA-pair) $h_{k} = [q, a_k]$ and predict their correctness scores based on the video and subtitle context~\cite{Onishi2016WhoDW}. 
We denote the ground-truth (GT) answer index as $y^\mathit{ans}$ and thus the GT hypothesis as $h_{y^\mathit{ans}}$.
We then extract video frames $\{v_t\}_{t=1}^{T}$ at 0.5 FPS ($T$ is the number of frames for each video). 
Subtitle sentences are then temporally aligned with the video frames.
Specifically, for each frame $v_t$, we pair it with two neighboring sentences based on the subtitle timestamps. 
We choose two neighbors since this keeps most of the sentences at our current frame rate, and also avoids severe misalignment between the frames and the sentences.
The set of aligned subtitle sentences are denoted as $\{s_t\}_{t=1}^{T}$. 
We denote the number of words in each hypothesis and subtitle as $L_h$, $L_s$, respectively. 
We use $N_o$ to denote the number of object regions in a frame, and $d=128$ as the hidden size. 

\subsection{STAGE Architecture}\label{sec:arch}
\paragraph{Input Embedding Layer} 
For each frame $v_t$, we use Faster R-CNN~\cite{Ren2015FasterRT} pre-trained on Visual Genome~\cite{krishna2017visual} to detect objects and extract their regional representation as our visual features~\cite{Anderson2017BottomUpAT}.
We keep the top-20 object proposals and use PCA to reduce the feature dimension from 2048 to 300, to save GPU memory and computation. 
We denote $o_{t,r} \in \mathbb{R}^{300}$ as the $\text{r}\mbox{-}th$ object embedding in the $\text{t}\mbox{-}th$ frame. 
To encode the text input, we use BERT~\cite{Devlin2018BERTPO}, a transformer-based language model~\cite{Vaswani2017AttentionIA} that achieves state-of-the-art performance on various NLP tasks. 
Specifically, we first fine-tune the BERT-base model using the masked language model and next sentence prediction objectives on the subtitles and QA pairs from TVQA+ train set. 
Then, we fix its parameters and use it to extract 768D word-level embeddings from the second-to-last layer for the subtitles and each hypothesis. 
Both embeddings are projected into a $128$D space using a linear layer with ReLU.

\paragraph{Convolutional Encoder}\label{sec:conv_enc}
Inspired by the recent trend of replacing recurrent networks with CNNs~\cite{Dauphin2016LanguageMW,Yu2018QANetCL} and Transformers~\cite{Vaswani2017AttentionIA,Devlin2018BERTPO} for sequence modeling, we use positional encoding (PE), CNNs, and layer normalization~\cite{Ba2016LayerN} to build our basic encoding block. 
As shown in the bottom-right corner of Fig.~\ref{fig:model_main}, it is comprised of a PE layer and multiple convolutional layers, each with a residual connection~\cite{He2016DeepRL} and layer normalization. 
We use $\mathrm{Layernorm}(\mathrm{ReLU}(\mathrm{Conv}(x)) + x)$ to denote a single $\mathrm{Conv}$ unit and stack $N_\mathrm{conv}$ of such units as the convolutional encoder. $x$ is the input after PE, $\mathrm{Conv}$ is a depthwise separable convolution~\cite{Chollet2017XceptionDL}.
We use two convolutional encoders at two different levels of STAGE, one with kernel size 7 to encode the raw inputs, and another with kernel size 5 to encode the fused video-text representation. 
For both encoders, we set $N_\mathrm{conv} =2$.

\paragraph{QA-Guided Attention} 
For each hypothesis $h_{k} = [q, a_k]$, we compute its attention scores w.r.t. the object embeddings in each frame and the words in each subtitle sentence, respectively. 
Given the encoded hypothesis $H_k \in \mathbb{R}^{L_h \times d}$ for the hypothesis $h_k$ with $L_h$ words, and encoded visual feature $V_t \in \mathbb{R}^{N_o \times d}$ for the frame $v_t$ with $N_o$ objects, we compute their matching scores $M_{k, t} \in \mathbb{R}^{L_h \times N_o} = H_k V_t^T$. 
We then apply softmax at the second dimension of $M_{k, t}$ to get the normalized scores $\bar{M}_{k, t}$. 
Finally, we compute the QA-aware visual representation $V_{k, t}^{att} \in \mathbb{R}^{L_h \times d} = \bar{M}_{k, t} V_t$. 
Similarly, we compute QA-aware subtitle representation $S_{k, t}^{att}$. 

\paragraph{Video-Text Fusion} 
The above two QA-aware representations are then fused together as:
\begin{align*}
    F_{k, t} = [S_{k, t}^{att}; V_{k, t}^{att}; S_{k, t}^{att} \odot V_{k, t}^{att}]W_F + b_F,
\end{align*}
where $\odot$ denotes hadamard product, $W_F \in \mathbb{R}^{3d \times d}$ and $b_F \in \mathbb{R}^{d}$ are trainable weights and bias, $F_{k, t} \in \mathbb{R}^{L_h \times d}$ is the fused video-text representation. 
After collecting $F_{k, t}^{att}$ from all time steps, we get $F_{k}^{att} \in \mathbb{R}^{T \times  L_h \times d}$. 
We then apply another convolutional encoder with a max-pooling layer to obtain the output $A_{k} \in \mathbb{R}^{T \times d}$.

\paragraph{Span Predictor} 
To predict temporal spans, we predict the probability of each position being the start or end of the span. 
Given the fused input $A_{k} \in \mathbb{R}^{T \times d}$, we produce start probabilities $\mathbf{p_{k}^1} \in \mathbb{R}^T$ and end probabilities $\mathbf{p_{k}^2} \in \mathbb{R}^T$  using two linear layers with softmax, as shown in the top-right corner of Fig.~\ref{fig:model_main}. Different from existing works~\cite{Seo2017BidirectionalAF,Yu2018QANetCL} that used the span predictor for text only, we use it for a joint localization of both video and text, which requires properly-aligned joint embeddings.

\paragraph{Span Proposal and Answer Prediction} 
Given the max-pooled video-text representation $A_{k}$, we use a linear layer to further encode it. 
We run max-pool across all the time steps to get a global hypothesis representation $G_{k}^{g} \in \mathbb{R}^d$. 
With the start and end probabilities from the span predictor, we generate span proposals using dynamic programming~\cite{Seo2017BidirectionalAF}. 
At training time, we combine the set of proposals with $\mathit{IoU} \geq 0.5$ with the GT spans, as well as the GT spans to form the final proposals $\{st_{p}, ed_{p}\}$~\cite{Ren2015FasterRT}. 
At inference time, we take the proposals with the highest confidence scores for each hypothesis. 
For each proposal, we generate a local representation $G_{k}^{l} \in \mathbb{R}^d$ by max-pooling $A_{k, st_p:ed_p}$. 
The local and global representations are concatenated to obtain $G_k \in \mathbb{R}^{2d}$. 
We then forward $\{G_k\}_{k=1}^5$ through softmax to get the answer scores $\mathbf{p}^\mathit{ans} \in \mathbb{R}^5$. 
Compared with existing works~\cite{Jang2017TGIFQATS,Zhao2017VideoQA} that use soft temporal attention, we use more interpretable hard attention, extracting local features (together with global features) for question answering.

\subsection{Training and Inference}\label{training_inference}
    
In this section, we describe the objective functions used in the STAGE framework. 
Since our spatial and temporal annotations are collected based on the question and GT answer, we only apply the attention loss and span loss on the targets associated with the GT hypothesis (question + GT answer), i.e., $M_{k=y^\mathit{ans},t}$, $\mathbf{p}^1_{k=y^\mathit{ans}}$ and $\mathbf{p}^2_{k=y^\mathit{ans}}$. 
For brevity, we omit the subscript $k\text{=}y^\mathit{ans}$ in the following. 

\paragraph{Spatial Supervision} 
While the attention described in Sec.~\ref{sec:arch} can be learned in a weakly supervised end-to-end manner, we can also train it with supervision from GT boxes.
We define a box as positive if it has an $\mathit{IoU} \geq 0.5$ with the GT box. 
Consider the attention scores $M_{t, j} \in \mathbb{R}^{N_o}$ from a concept word $w_j$ in GT hypothesis $h_{y^\mathit{ans}}$ to the set of proposal boxes' representations $\{o_{t, r}\}_{r=1}^{N_o}$ at frame $v_t$. 
We expect the attention on positive boxes to be higher than the negative ones, and therefore use LSE~\cite{Li2017ImprovingPR} loss for the supervision:

\begin{align*}
    \small{\mathcal{L}_{t, j}} \mbox{=} \sum_{r_p \in \Omega_{p}, r_n \in \Omega_{n}}\log \Big(1 + \exp(M_{t,j,r_n} - M_{t,j,r_p})\Big),
\end{align*}

\noindent where $M_{t,j,r_p}$ is the $\text{r}_p\mbox{-}th$ element of the vector $M_{t, j}$. $\Omega_{p}$ and $\Omega_{n}$ denote the set of positive and negative box indices, respectively. LSE loss is a smoothed alternative to the widely used hinge loss, it is easier to optimize than the original hinge loss~\cite{Li2017ImprovingPR}.
During training, we randomly sample two negatives for each positive box. 
We use $\mathcal{L}^{att}_i$ to denote the attention loss for the $\text{i}\mbox{-}th$ example, which is obtained by summing over all the annotated frames $\{v_t\}$ and concepts $\{w_j\}$ for $\mathcal{L}^{att}_{t,j}$. 
We define the overall attention loss $\mathcal{L}^{att} = \frac{1}{N} \sum_{i=1}^{N} \mathcal{L}^{att}_i$.
At inference time, we choose the boxes with scores higher than 0.2 as the predictions.

\paragraph{Temporal Supervision} 
Given softmax normalized start and end probabilities $\mathbf{p}^{1}$ and $\mathbf{p}^{2}$, we apply cross-entropy loss:
\begin{align*}
    \mathcal{L}^\mathit{span} = - \frac{1}{2N} \sum_{i=1}^{N}\big(\log \mathbf{p}^1_{y_i^1} + \log \mathbf{p}^2_{y_i^2} \big),
\end{align*}
\noindent where $y_i^1$ and $y_i^2$ are the GT start and end indices. 

\paragraph{Answer Prediction} 
Similarly, given answer probabilities $\mathbf{p}^{ans}$, our answer prediction loss is:
\begin{align*}
    \mathcal{L}^\mathit{ans} = - \frac{1}{N} \sum_{i=1}^{N} \log \mathbf{p}^\mathit{ans}_{y_i^\mathit{ans}},
\end{align*}
\noindent where $y_i^\mathit{ans}$ is the index of the GT answer. 

Finally, the overall loss is a weighted combination of the three objectives above: $\mathcal{L} = \mathcal{L}^\mathit{ans} + w_{att} \mathcal{L}^{att} + w_\mathit{span} \mathcal{L}^\mathit{span}$, where $w_{att}$ and $w_\mathit{span}$ are set as $0.1$ and $0.5$ based on validation set tuning.

%%%%%%%%%%%%%%%%%%%%%%%%%%%%%%%%%%%%%%
% Section 5 Experiments
%%%%%%%%%%%%%%%%%%%%%%%%%%%%%%%%%%%%%%
\section{Experiments}\label{exp}
As introduced, our task is spatio-temporal video question answering, requiring systems to temporally localize relevant moments, spatially detect referred objects and people, and answer questions. 
In this section, we first define the evaluation metrics, then compare STAGE against several baselines, and finally provide a comprehensive analysis of our model. 
Additionally, we also evaluate STAGE on the full TVQA dataset.

\subsection{Metrics}

To measure QA performance, we use classification accuracy (QA Acc.). 
We evaluate \textit{span prediction} using temporal mean Intersection-over-Union (Temp. mIoU) following previous work~\cite{Hendricks2017LocalizingMI} on language-guided video moment retrieval. 
Since the span depends on the hypothesis (QA pair), each QA pair provides a predicted span, but we only evaluate the span of the predicted answer.
Additionally, we propose Answer-Span joint Accuracy (ASA), that jointly evaluates both answer prediction and span prediction. 
For this metric, we define a prediction to be correct if the predicted span has an $\mathit{IoU} \geq 0.5$ with the  GT span, provided that the answer prediction is correct. 
Finally, to evaluate \textit{object grounding} performance, we follow the standard metric from the PASCAL VOC challenge~\cite{everingham2015pascal} and report the mean Average Precision (Grd. mAP) at $\mathit{IoU}$ threshold 0.5. We only consider the annotated words and frames when calculating the mAP.

\begin{table}[!t]
\centering
\small
\scalebox{0.85}{
\begin{tabular}{lcccc}
\toprule
\multirow{2}{*}{Model}  & QA & Grd.  & Temp. & \multirow{2}{*}{ASA} \\
&  Acc. & mAP & mIoU &  \\
\midrule
ST-VQA~\cite{Jang2017TGIFQATS}  & 48.28 & - & - & - \\ 
two-stream~\cite{lei2018tvqa}  & 68.13 & - & - & - \\ 
STAGE (video) & 52.75 & 26.28 & 10.90 & 2.76 \\ 
STAGE (sub) & 67.99 & - & 30.16 & 20.13 \\ 
STAGE & \textbf{74.83} & \textbf{27.34} & \textbf{32.49} & \textbf{22.23} \\ 
\midrule
Human~\cite{lei2018tvqa} & 90.46 & - & - & - \\
\bottomrule
\end{tabular}
}
\caption{TVQA+ test set results.}
\label{tab:main_res}
\end{table}

\subsection{Comparison with Baseline Methods}
We consider the two-stream model~\cite{lei2018tvqa} as our main baseline. 
In this model, two streams are used to predict answer scores from subtitles and videos respectively and final answer scores are produced by summing scores from both streams. 
We retrain the model using the official code\footnote{\url{https://github.com/jayleicn/TVQA}} on TVQA+ data, with the same feature as STAGE. 
We also consider ST-VQA~\cite{Jang2017TGIFQATS} model, which is primarily designed for question answering on short videos (GIFs).
We also provide STAGE variants that use only video or subtitle to study the effect of using only one of the modalities.
Table~\ref{tab:main_res} shows the test results of STAGE and the baselines. STAGE outperforms the baseline model (two-stream) by a large margin in QA Acc.,\footnote{This also holds true when considering mean (standard-deviation) of 5 runs: 74.20 (0.42).} with 9.83\% relative gains.
Additionally, STAGE also localizes the relevant moments with temporal mIoU of 32.49\% and detects referred objects and people with mAP of 27.34\%. 
However, a large gap is still observed between STAGE and human, showing space for further improvement.

\begin{table}[!t]
\centering
\small
\scalebox{0.84}{
\begin{tabular}{lcccccc}
\toprule
\multirow{2}{*}{Model}  & QA & Grd.  & Temp. & \multirow{2}{*}{ASA} \\
&  Acc. & mAP & mIoU &  \\
\midrule
baseline & 65.79 & 2.74 & - & - \\
+ CNN  & 67.25 & 3.16 & - & - \\
+ Aligned Fusion (backbone)  & 68.31 & 7.31 & - & - \\
+ Temp. Sup.  & 71.40 & 10.86 & 30.77 & 20.09 \\
+ Spat. Sup.  & 71.99 & 24.10 & 31.16 & 20.42 \\
+ Local Feature (STAGE)   & \textbf{72.56} & \textbf{25.22} & \textbf{31.67} & \textbf{20.78} \\ 
\midrule
STAGE with GT Span & 73.28 & - & - & -\\ 
\bottomrule
\end{tabular}
}
\caption{Ablation study of STAGE on TVQA+ val set. \textit{Each row adds an extra component to the row above it.}}
\label{tab:ablation}
\end{table}

\begin{figure*}[!ht]
  \centering
  \scalebox{0.9}{
    \includegraphics[width=\linewidth]{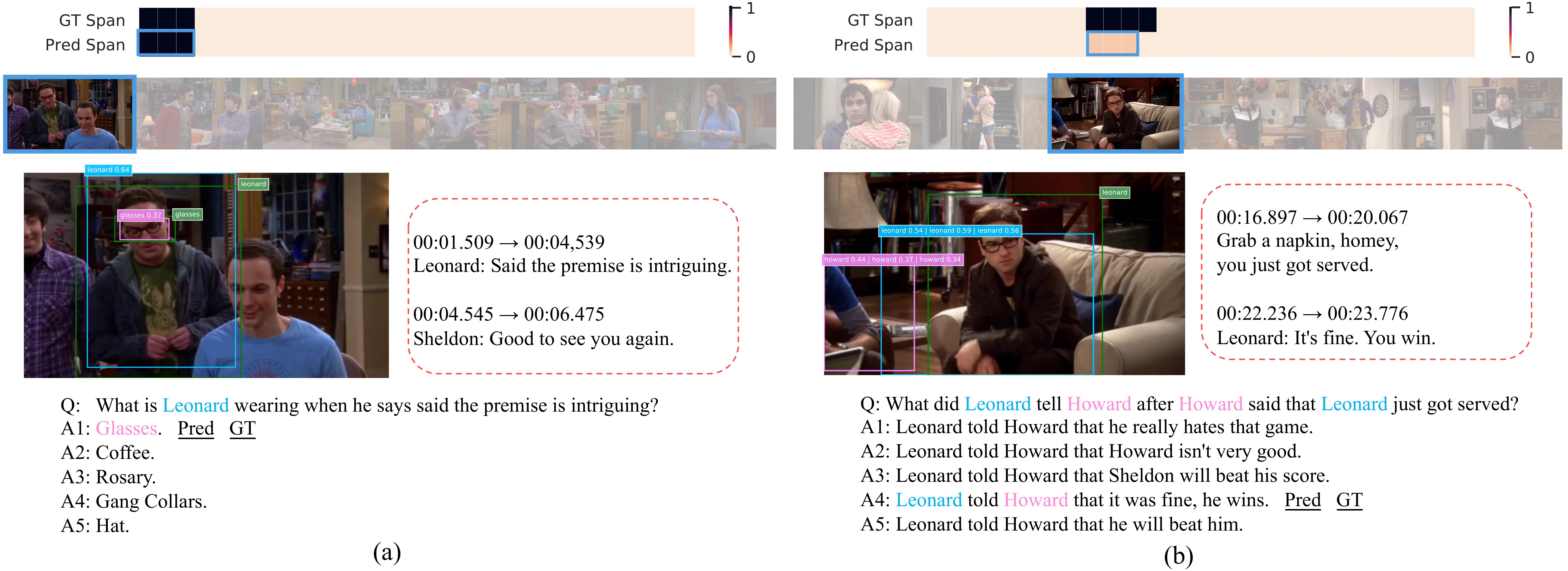}
  }
  \caption{Example predictions from STAGE. Span predictions are shown on the top, each block represents a frame, the color indicates the model's confidence for the spans. For each QA, we show grounding examples and scores for one frame in GT span. GT boxes are in green. Predicted and GT answers are labeled by \underline{Pred} and \underline{GT}, respectively.}
  \label{fig:correct_example}
\end{figure*}

\begin{table}[!t]
\centering
\small
\scalebox{0.79}{
\begin{tabular}{ccccccc}
\toprule
Model & baseline & +CNN & +AF & +TS & +SS & +LF \\ 
\midrule
what (60.52\%) & 65.66 & 66.43  & 67.58 & 70.76 & 71.25 & \textbf{72.34} \\
who (10.24\%) & 65.37  & 64.08 & 64.72  & 72.17 & 73.14 & \textbf{74.11} \\
where (9.68\%) & 65.41 & 64.38 & 68.49 & 71.58 & 71.58 & \textbf{74.32} \\
why (9.55\%) & 74.31  & 78.82 & 77.43  & \textbf{79.86} & 78.12 & 76.39 \\
how (9.05\%) & 60.81  & 67.03 & 69.23  & 66.30 & \textbf{69.96} & 67.03 \\ 
\midrule
total (100\%) & 65.79 & 67.25 & 68.31  & 71.40 & 71.99 & \textbf{72.56} \\ 
\bottomrule
\end{tabular}
}
\caption{QA Acc. by question type on TVQA+ val set. For brevity, we only show top-5 question types (percentage in brackets). 
AF=Aligned Fusion, TS=Temp. Sup., SS=Spat. Sup., LF=Local Feature. \textit{Each column adds an extra component to the column before it.}
}
\label{tab:acc_by_qtype}
\end{table}

\subsection{Model Analysis}
\paragraph{Backbone Model} 
Given the full STAGE model defined in Sec.~\ref{sec:methods}, we define the \textit{backbone model} as the ablated version of it, where we remove the span predictor along with the span proposal module, as well as the explicit attention supervision.
We further replace the CNN encoders with RNN encoders, and remove the aligned fusion from the backbone model. 
This baseline model uses RNN to encode input sequences and interacts QA pairs with subtitles and videos separately. The final confidence score is the sum of the confidence scores from the two modalities. 
In the backbone model, we align subtitles with video frames from the start, fusing their representation conditioned on the input QA pair, as in Fig.~\ref{fig:model_main}.
We believe this aligned fusion is essential for improving QA performance, as the latter part of STAGE has a joint understanding of both video and subtitles. 
With both changes, our backbone model obtains 68.31\% on QA Acc., significantly higher than the baseline's 65.79\%. The results are shown in Table~\ref{tab:ablation}.

\paragraph{Temporal and Spatial Supervision} 
In Table~\ref{tab:ablation}, we also show the results when using temporal and spatial supervision. After adding temporal supervision, the model is be able to ground on the temporal axis, which also improves the model's performance on other tasks. Adding spatial supervision gives additional improvements, particularly for Grd. mAP, with 121.92\% relative gain.

\paragraph{Span Proposal and Local Feature} 
In the second-to-last row of Table~\ref{tab:ablation}, we show our full STAGE model, which is augmented with local features $G^l$ for question answering. 
Local features are obtained by max-pooling the span proposal regions, which contain more relevant cues for answering the questions. 
With $G^l$, we achieve the best performance across all metrics, indicating the benefit of using local features. 

\paragraph{Inference with GT Span} 
The last row of Table~\ref{tab:ablation} shows our model uses GT spans instead of predicted spans at inference time. We observe better QA Acc. with GT spans.

\paragraph{Accuracy by Question Type} 
In Table~\ref{tab:acc_by_qtype}, we show a breakdown of QA Acc. by question type. 
We observe a clear increasing trend on ``what", ``who", and ``where" questions after using the backbone net and adding attention/span modules in each column.  
Interestingly, for ``why" and ``how" questions, our full model fails to present overwhelming performance, indicating some reasoning (textual) module to be incorporated as future work.

\paragraph{Qualitative Examples} 
We show two correct predictions in Fig.~\ref{fig:correct_example}, where Fig.~\ref{fig:correct_example}(a) uses grounded objects to answer the question, and Fig.~\ref{fig:correct_example}(b) uses text. More examples (including failure cases) are provided in the appendix.

\begin{table}[!t]
\centering
\small
\scalebox{0.72}{
\begin{tabular}{lccc}
\toprule
Model & Temp. Sup. & val & test-public \\ 
\midrule
two-stream~\cite{lei2018tvqa} & \xmark & 65.85 & 66.46 \\
PAMN~\cite{Kim2019ProgressiveAM} & \xmark & 66.38 & 66.77 \\ 
multi-task~\cite{Kim2019GainingES} & \cmark & 66.22 & 67.05 \\
\midrule
STAGE backbone (GloVe)  & \xmark  & 66.46 & - \\
STAGE backbone + Temp. Sup. (GloVe)  & \cmark  & 66.92 & - \\
STAGE backbone  & \xmark  & 68.56 & 69.67 \\
STAGE backbone + Temp. Sup. & \cmark &\textbf{70.50} & \textbf{70.23} \\
\bottomrule
\end{tabular}
}
\caption{QA Acc. on the full TVQA dataset.}
\label{tab:full_dataset_res}
\end{table}

\paragraph{TVQA Results} 
We also conduct experiments on the full TVQA dataset (Table~\ref{tab:full_dataset_res}), without relying on the bounding boxes and refined timestamps in TVQA+.
Without temporal supervision, STAGE backbone is able to achieve 3.91\% relative gain from the best published result (multi-task) on TVQA test-public set. 
Adding temporal supervision, performance is improved to 70.23\%. For a fair comparison, we also provided STAGE variants using GloVe~\cite{pennington2014glove} instead of BERT~\cite{Devlin2018BERTPO} as text feature. Using GloVe, STAGE models still achieve better results.

%%%%%%%%%%%%%%%%%%%%%%%%%%%%%%%%%%%%%%
% Section 6 Conclusion
%%%%%%%%%%%%%%%%%%%%%%%%%%%%%%%%%%%%%%
\section{Conclusion}\label{conclusion}

We collected the TVQA+ dataset and proposed the spatio-temporal video QA task. 
This task requires systems to jointly localize relevant moments, detect referred objects/people, and answer questions. 
We further introduced STAGE, an end-to-end trainable framework to jointly perform all three tasks.
Comprehensive experiments show that temporal and spatial predictions help improve QA performance, as well as providing explainable results. 
Though our STAGE achieves state-of-the-art performance, there is still a large gap compared with human performance, leaving space for further improvement.

%%%%%%%%%%%%%%%%%%%%%%%%%%%%%%%%%%%%%%
% Section Acknowledgement
%%%%%%%%%%%%%%%%%%%%%%%%%%%%%%%%%%%%%%
\section*{Acknowledgement}
We thank the reviewers for their helpful feedback. This research is supported by NSF Awards \#1633295, 1562098, 1405822, DARPA MCS Grant \#N66001-19-2-4031, DARPA KAIROS Grant \#FA8750-19-2-1004, Google Focused Research Award, and ARO-YIP Award \#W911NF-18-1-0336.

\bibliography{acl2020}
\bibliographystyle{acl_natbib}

%%%%%%%%%%%%%%%%%%%%%%%%%%%%%%%%%%%%%%
% Section Appendix
%%%%%%%%%%%%%%%%%%%%%%%%%%%%%%%%%%%%%%
\appendix
\section{Appendices}\label{sec:appendix}

\subsection{Timestamp Annotation}
\begin{figure}[!ht]
    \centering
  \includegraphics[width=\linewidth]{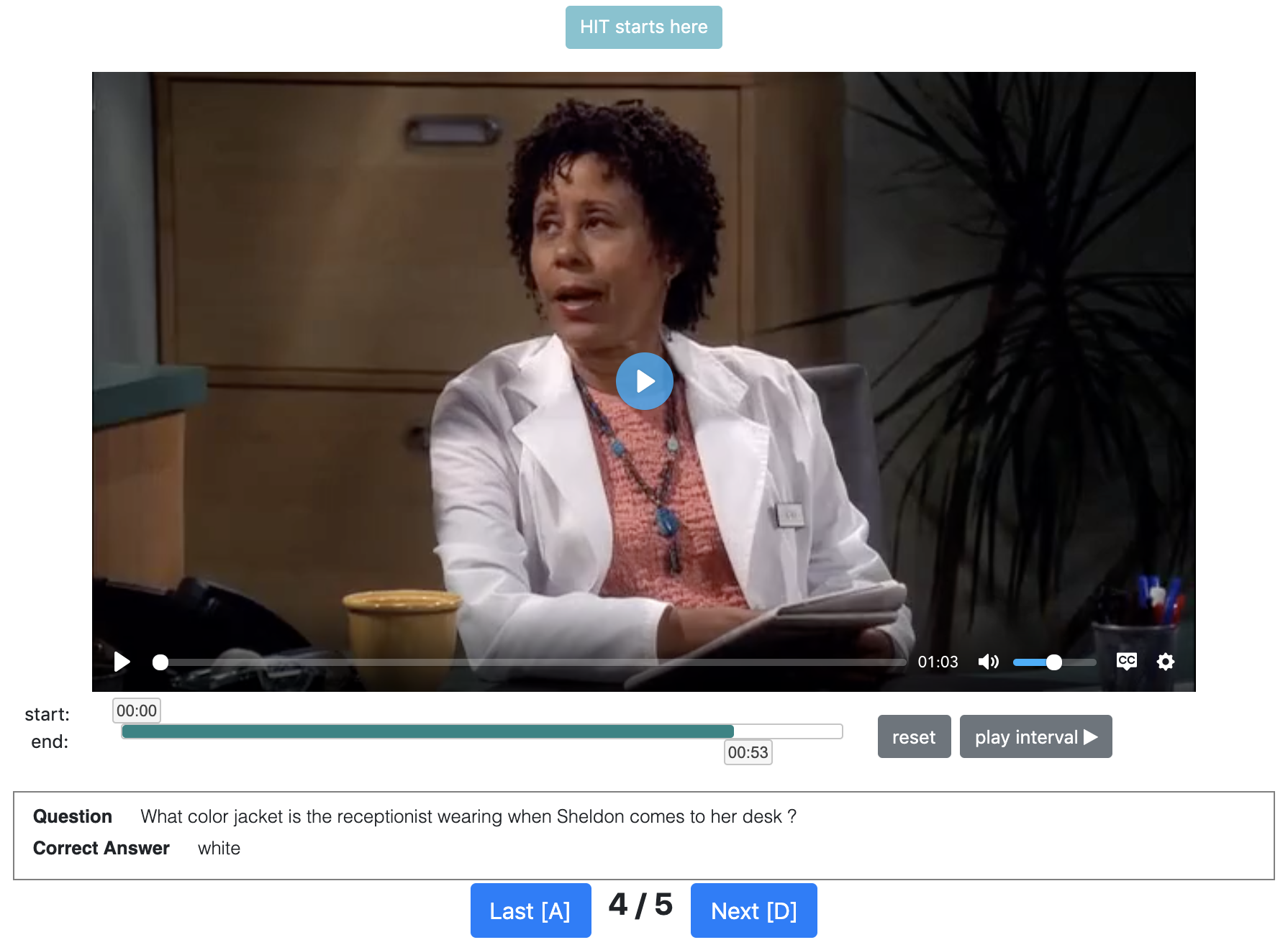}
  \caption{Timestamp refinement interface.}
  \label{fig:timestamp_interface}
\end{figure}

During our initial analysis, we find the original timestamp annotations from the TVQA~\cite{lei2018tvqa} dataset to be somewhat loose, i.e., around 8.7\% of 150 randomly sampled training questions had a span that was at least 5 seconds longer than what is needed. To have better timestamps, we asked a set of Amazon Mechanical Turk (AMT) workers to refine the original timestamps. Specifically, we take the questions that have a localized span length of more than 10 seconds (41.33\% of the questions) for refinement while leaving the rest unchanged. During annotation, we show a question, its correct answer, its associated video (with subtitle), as well as the original timestamp to the AMT workers (illustrated in Fig.~\ref{fig:timestamp_interface}, with instructions omitted). The workers are asked to adjust the start and end timestamps to make the span as small as possible, but need to contain all the information mentioned in the QA pair.

\begin{figure}[ht]
  \centering
  \includegraphics[width=\linewidth]{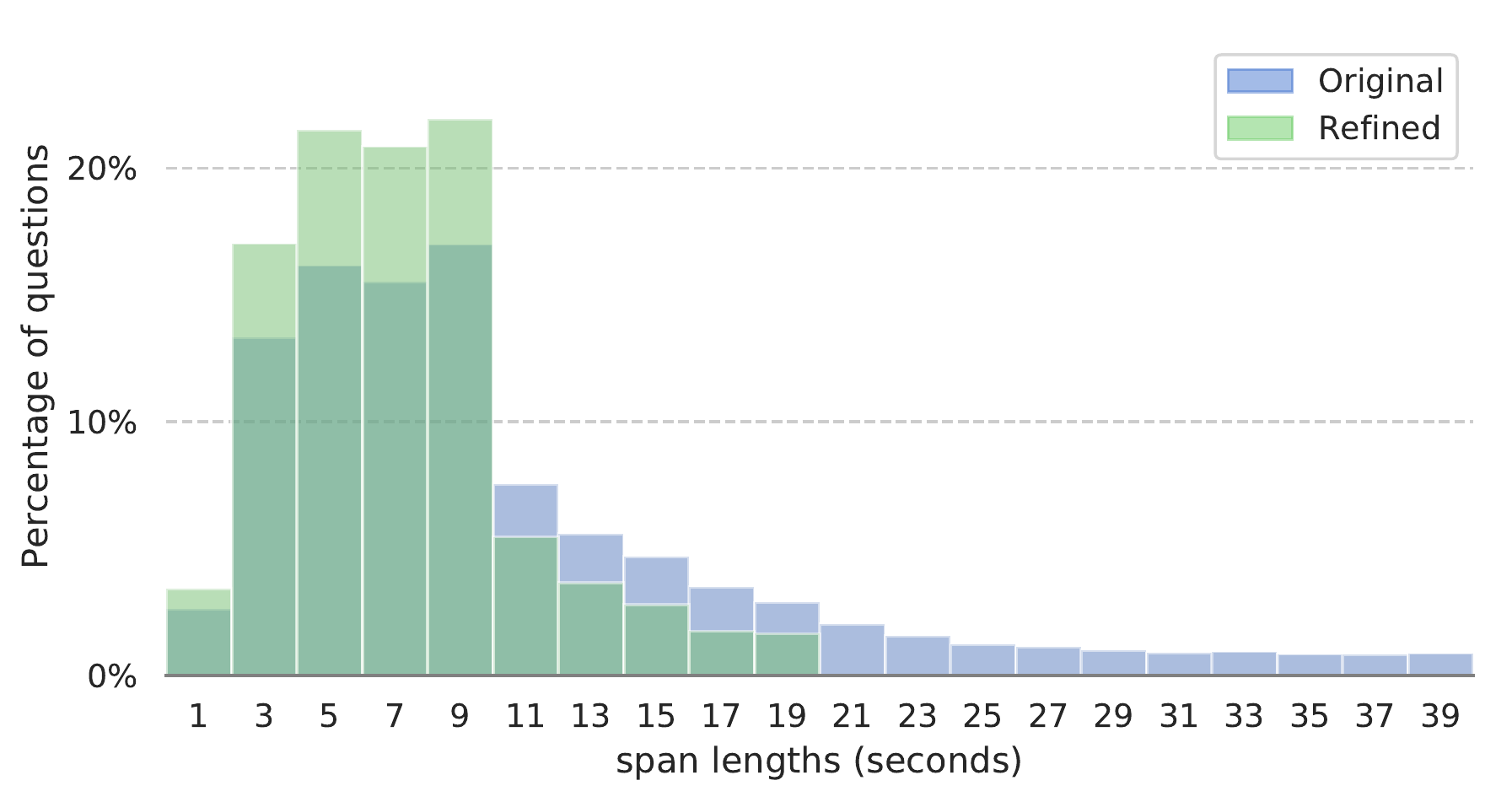}
  \caption{Comparison between the original and the refined timestamps in the TVQA+ train set. The refined timestamps are generally tighter than the original ones.}
  \label{fig:original_old_ts}
\end{figure}

We show span length distributions of the original and the refined timestamps from TVQA+ train set in Fig.~\ref{fig:original_old_ts}. The average span length of the original timestamps is 14.41 secs, while the average for the refined timestamps is 7.2 secs. 

In Table~\ref{tab:ts_performance_comparison} we show STAGE performance on TVQA+ val set using the original timestamps and the refined timestamps. Models with the refined timestamps performs consistently better than the ones with the original timestamps.

\begin{table}[ht]
\centering
\small
\scalebox{1.0}{
\begin{tabular}{lcc}
\hline
\multirow{2}{*}{Model} & \multicolumn{2}{c}{QA Acc.} \\ \cmidrule(rl){2-3} 
  & Original & Refined \\ \hline
 STAGE backbone & 68.31 & 68.31 \\
 + Temp. Sup. & 70.87 & 71.40 \\
 + Spat. Sup. & 71.23 & 71.99 \\
 + Local Feature (STAGE) & 70.63 & \textbf{72.56} \\ \hline
\end{tabular}
}
\caption{STAGE performance comparison between the original timestamps and the refined timestamps on TVQA+ val set. \textit{Each row adds an extra component to the row above it.}}
\label{tab:ts_performance_comparison}
\end{table}

\begin{figure}[t]
  \centering
  \includegraphics[width=0.94\linewidth]{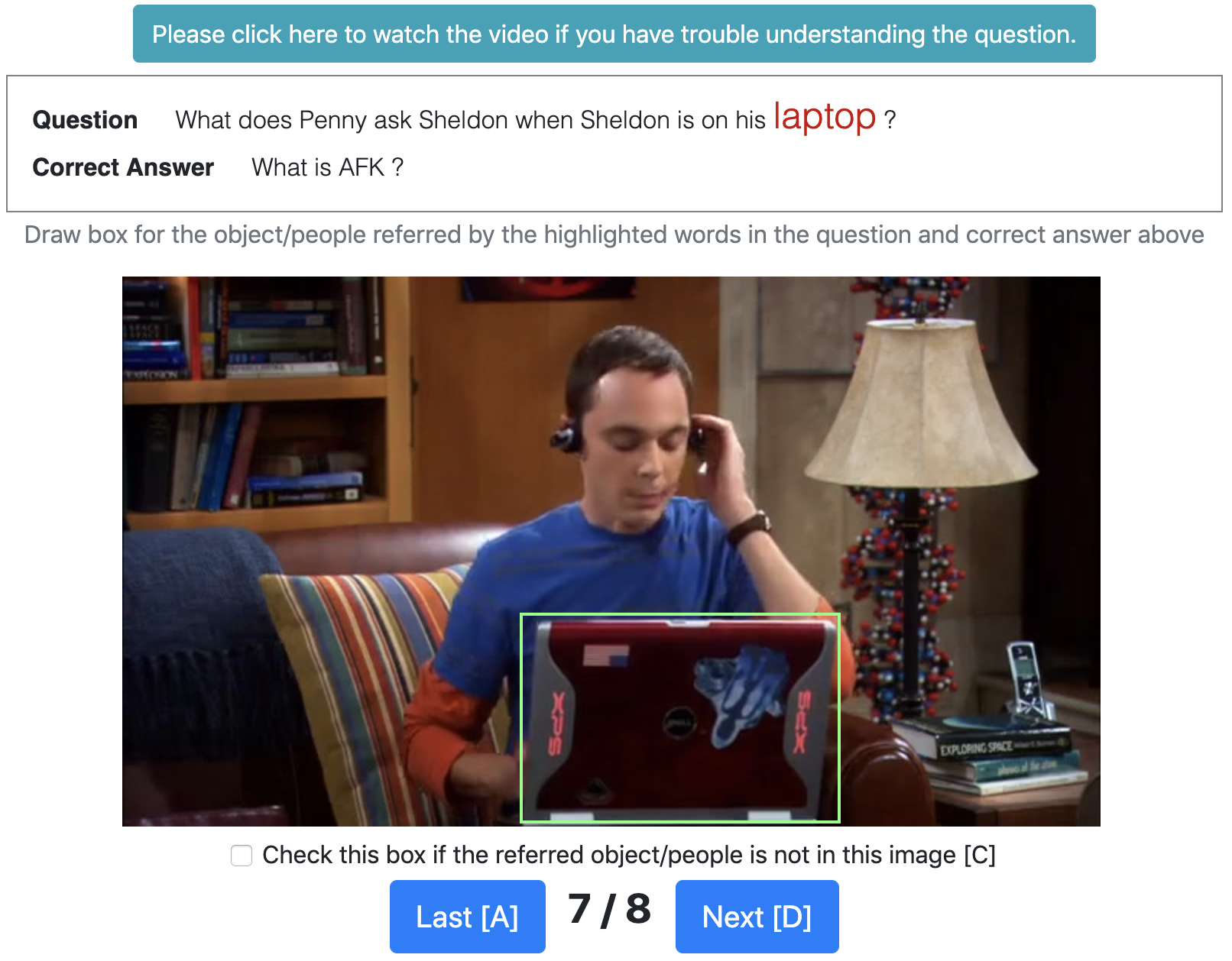}
  \caption{Bounding box annotation interface. Here, the worker is asked to draw a box around the highlighted word ``laptop''.}
  \label{fig:box_anno}
\end{figure} 

\subsection{Bounding Box Annotation}
At each step, we show a question, its correct answer, and the sampled video frames to an AMT worker. (illustrated in Fig.~\ref{fig:box_anno}). We do not annotate the wrong answers as most of them cannot be grounded in the video. We checked 200 sampled QAs - only 3.13\% of the wrong answers could be grounded, while 46\% of the correct answers could be grounded.
As each QA pair has multiple visual concepts as well as multiple frames, each task shows one pair of a concept word and a sampled frame.
For example, in Fig.~\ref{fig:box_anno}, the word ``laptop" is highlighted, and workers are instructed to draw a box around it. 
In our MTurk instructions, we required workers to draw boxes for each instance of a plural word. 
E.g., for the word ``everyone'', the worker need to draw a box for each person in the frame.
Note, it is possible that the highlighted word will be a non-visual word or a visual word that is not present in the frame being shown. 
In that case, the workers are allowed to check the box indicating the object is not present. 
Recent works~\cite{Zellers2018FromRT,Gu2018AVAAV} suggest the use of pre-trained detectors for semi-automated annotation. 
However, since TVQA+ has a wide range of categories (see Fig.~\ref{fig:cat_dist} and Table~\ref{tab:dataset_comparison}), it is challenging to use off-the-shelf detectors in the annotation process. 
As face detection and recognition might be easier than recognizing open set objects, we initially also tried using strong face detection~\cite{zhang2016joint} and recognition~\cite{liu2017sphereface} model for character face annotation, but the quality was much poorer than expected. 
Thus, we decided to invest the required funds to collect boxes manually and ensure their accuracy. 
After the collection, with the GT labels, we again used the above models to test face retrieval performance for 12 most frequently appeared characters in TVQA+. 
To allow \cite{liu2017sphereface} to work, we manually collected 5 GT faces for each character as our gallery set. 
At test time, we assign each test face the label of its closest neighbor from the gallery set in the learned embedding space. This method achieves 55.6 F1/74.4 Precision/44.4 Recall. 
Such performance is not strong enough to support further research. 
We found the main reason is due to many partial occlusion of faces (e.g., side faces) in TV shows. 

\subsection{Quality}
To ensure the quality of the collected bounding boxes, we only allow workers from English-speaking countries to participate the task. Besides, we set high requirements for workers -- they needed to have at least 3000 accepted HITs and 95\% accept rate. Qualified workers were well paid. We also kept track of the quality of the data during collection - workers with poor annotations were disqualified to work on our task. 
After collection, we further conducted an in-house check, 95.5\% of 200 sampled QAs are correctly labeled, indicating the high quality of our data.

\subsection{Training Details}
We optimize our model using Adam with an initial learning rate of 1e-3, weight decay 3e-7.
A mini-batch contains 16 questions. We train the model for maximum 100 epochs with early stop -- if QA Acc. is not improving for consecutive 5 epochs, the training is stopped. CNN hidden size is set to 128.

\begin{table}[t]
\centering
\small
\scalebox{1.0}{
\begin{tabular}{lcccc}
\toprule
\multirow{2}{*}{Model}  & QA & Grd.  & Temp. & \multirow{2}{*}{ASA} \\
&  Acc. & mAP & mIoU &  \\
\midrule
STAGE-LXMERT & 71.46 & 21.01 & 26.31 & 18.04 \\ 
STAGE & \textbf{74.83} & \textbf{27.34} & \textbf{32.49} & \textbf{22.23} \\ 
\bottomrule
\end{tabular}
}
\caption{TVQA+ test set results with LXMERT.}
\label{tab:res_with_lxmert}
\end{table}

\subsection{Vision-Language Pretrained Features}
In addition, we also consider features from LXMERT~\cite{tan2019lxmert}. 
This model is pretrained on a large amount of image-text pairs from multiple image captioning~\cite{lin2014microsoft,krishna2017visual} and image question answering~\cite{goyal2017making,hudson2019gqa,zhu2016cvpr} datasets. 
Specifically, we use video frame-question pairs as input to LXMERT, and use the extracted features to replace Faster R-CNN object features and BERT question features.
For answers and subtitles, we still use the original BERT features. 
The results are shown in Table~\ref{tab:res_with_lxmert}.
We notice that using LXMERT feature lowers STAGE's performance.
This is not surprising, as the domains in which the LXMERT model are pre-trained on are very different from TVQA+: (captions/questions+image) vs (subtitles+QAs+videos). Future work includes more investigation into adapting these pre-trained vision-language models for more challenging video+dialogue domains.

\begin{figure*}[t]
  \includegraphics[width=\linewidth]{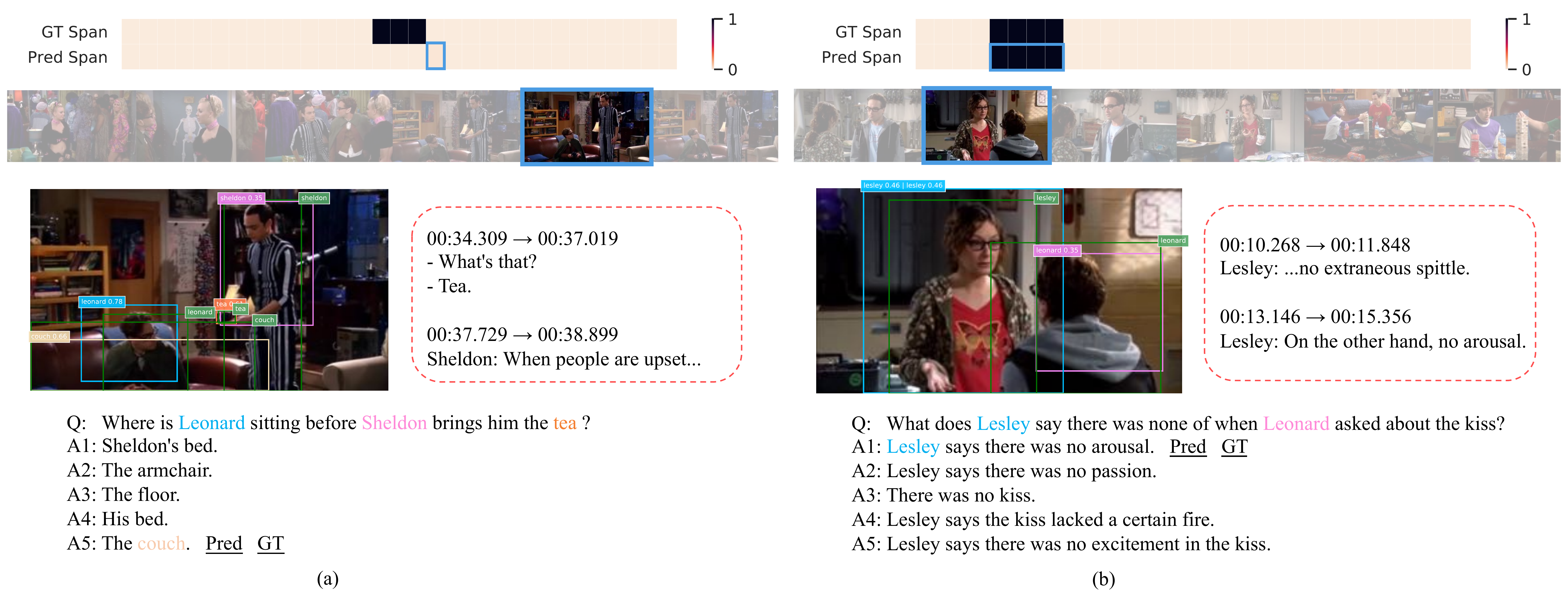}
  \includegraphics[width=\linewidth]{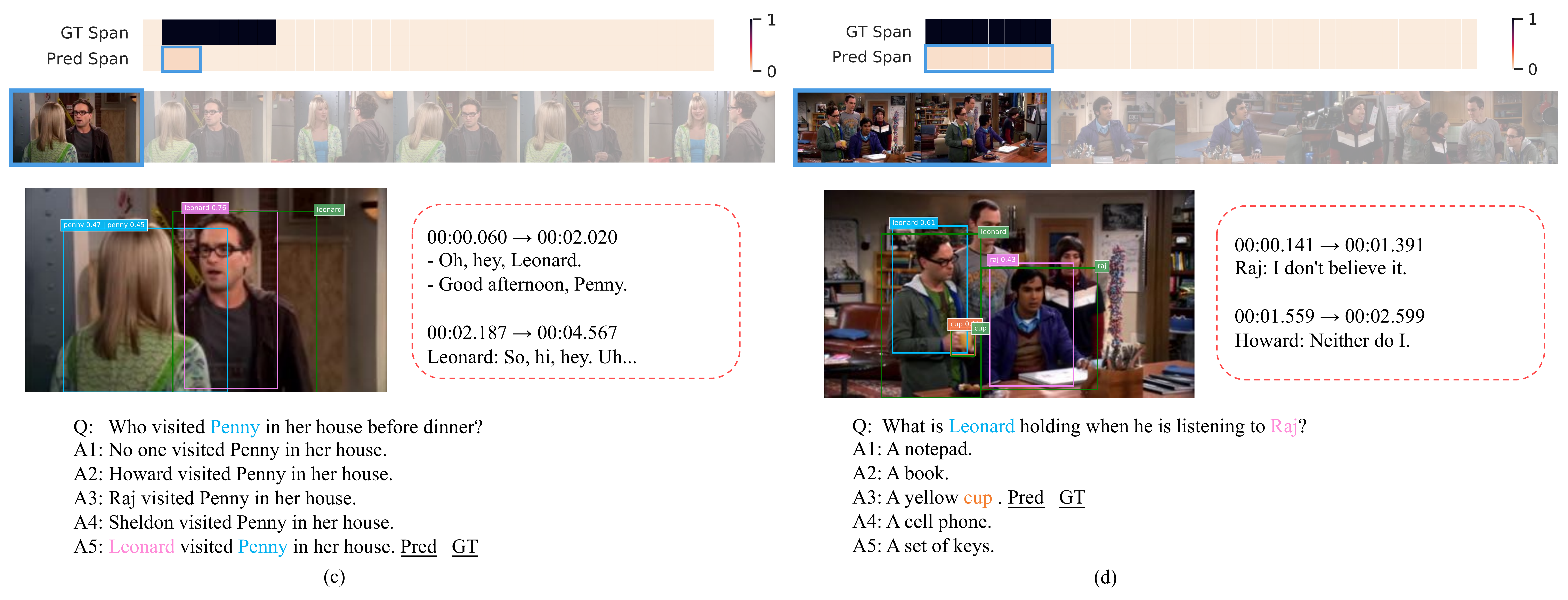}
  \includegraphics[width=\linewidth]{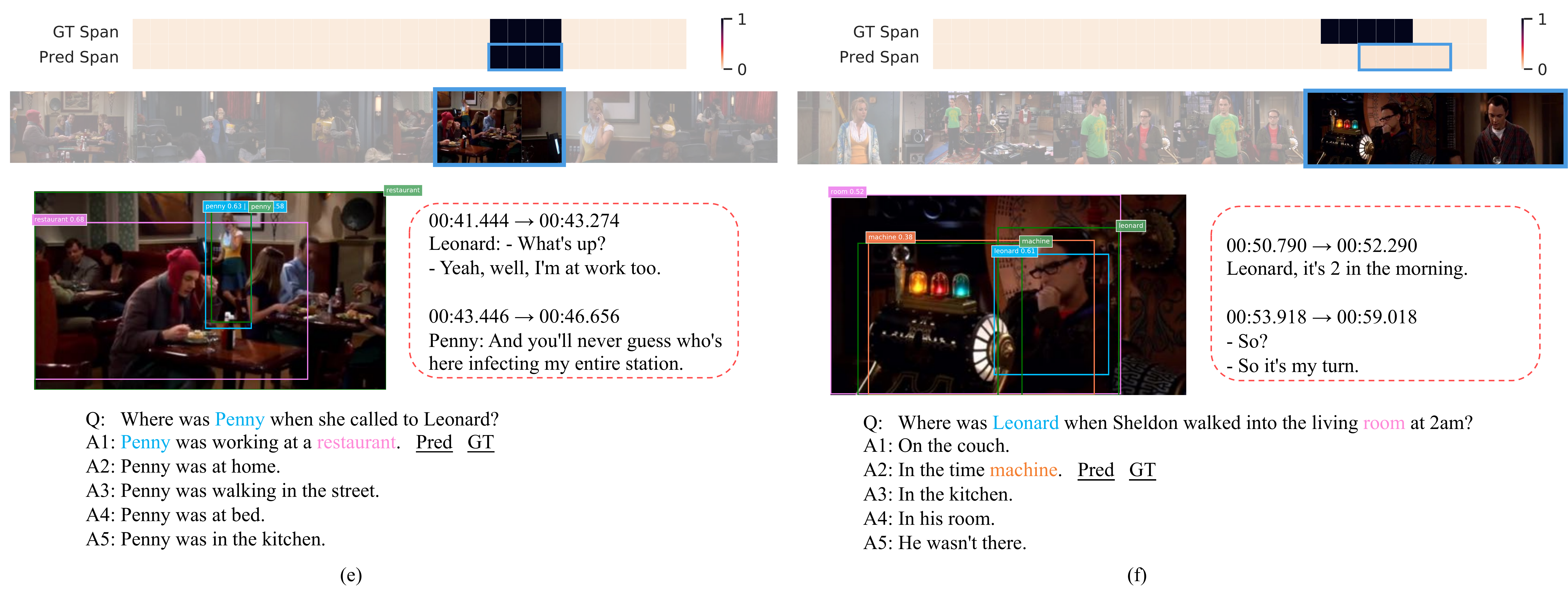}
  \caption{Correct prediction examples from STAGE. The span predictions are shown on the top of each example, each block represents a frame, the color indicates the model's confidence for the predicted spans. For each QA, we show grounding examples and scores for one frame in GT span, GT boxes are shown in green. Model predicted answers are labeled by \underline{Pred}, GT answers are labeled by \underline{GT}.}
  \label{fig:correct_examples}
\end{figure*}

\begin{figure*}[t]
  \includegraphics[width=\linewidth]{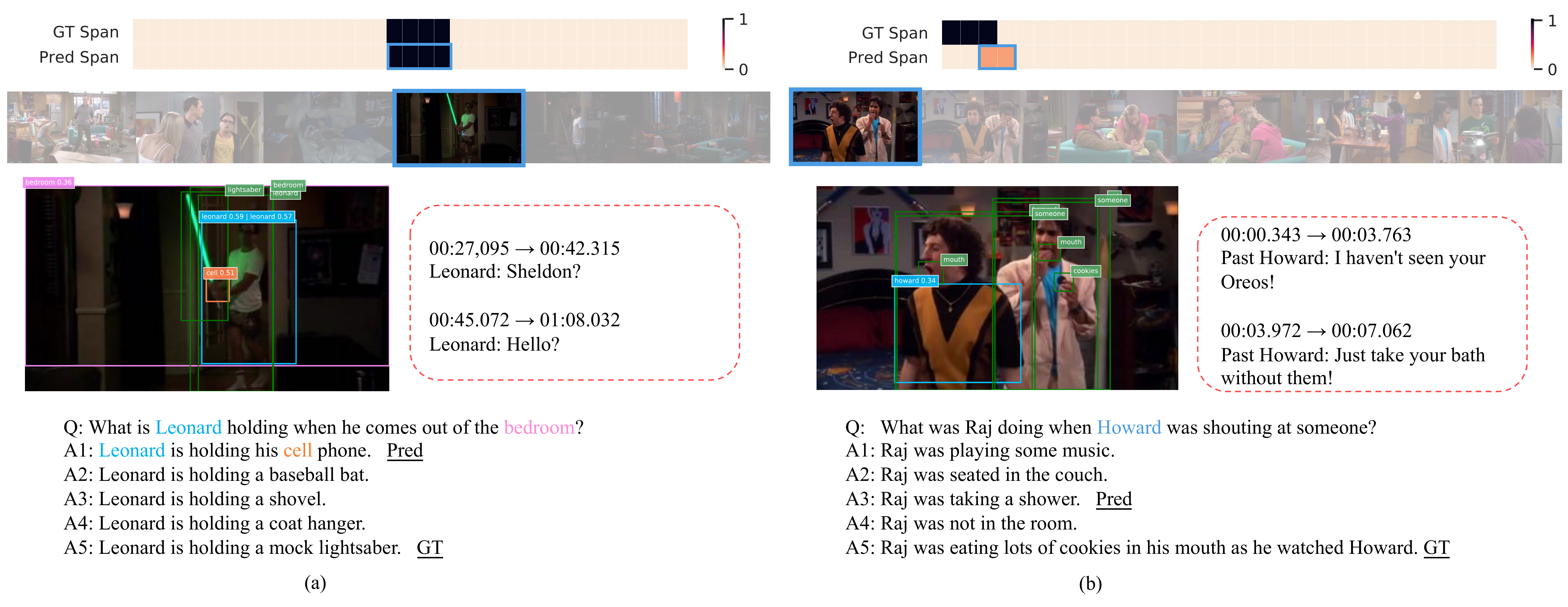}
  \includegraphics[width=\linewidth]{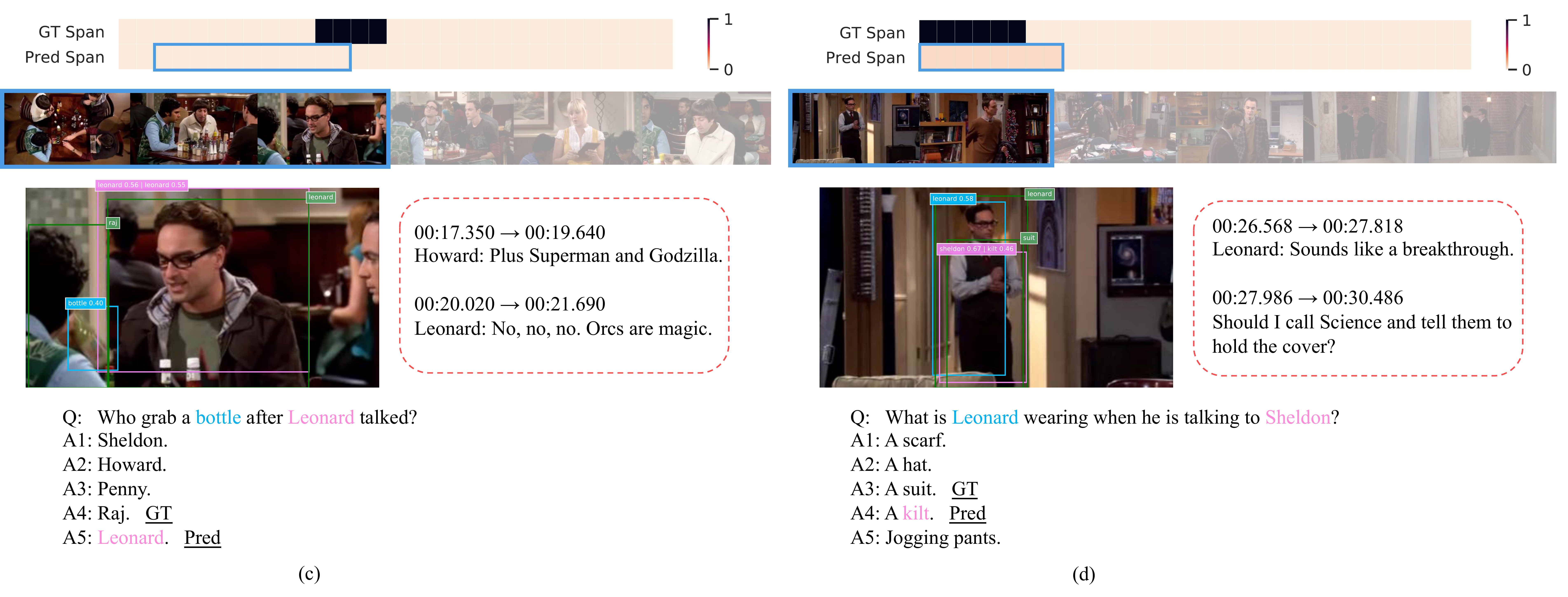}
  \includegraphics[width=\linewidth]{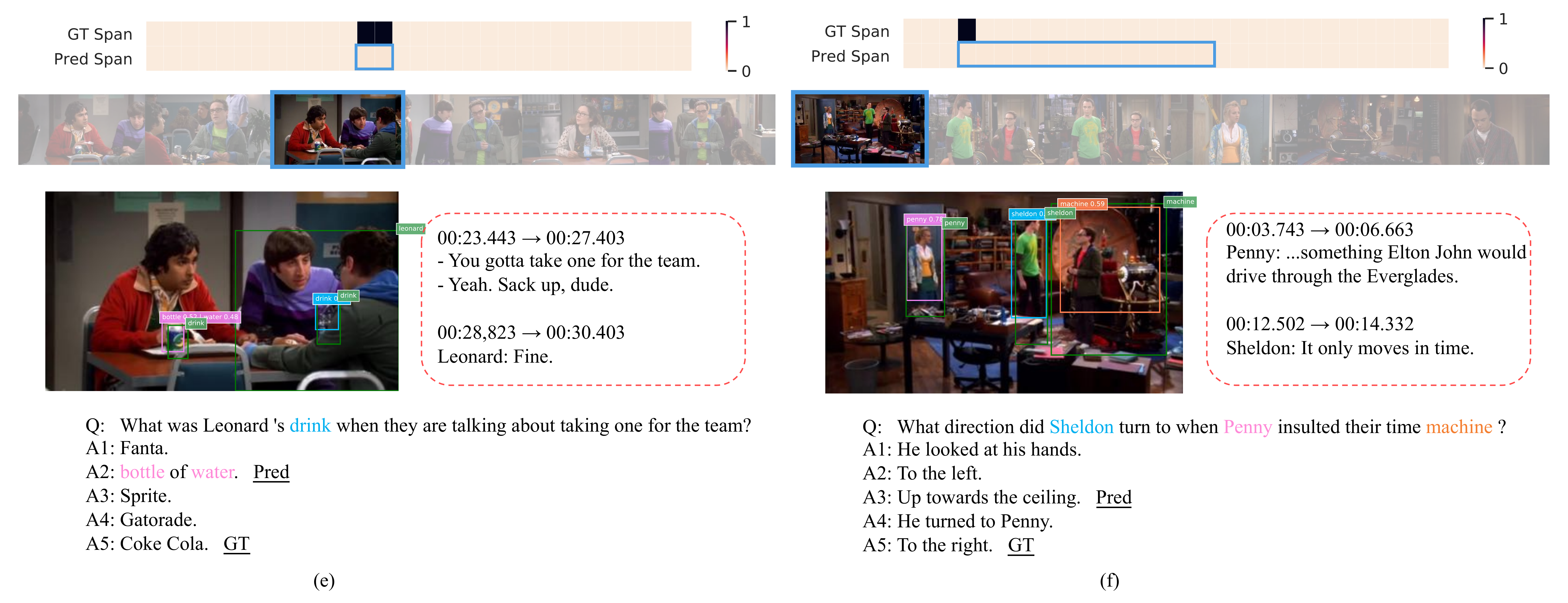}
  \caption{Wrong prediction examples from STAGE. The span predictions are shown on the top of each example, each block represents a frame, the color indicates the model's confidence for the predicted spans. For each QA, we show grounding examples and scores for one frame in GT span, GT boxes are shown in green. Model predicted answers are labeled by \underline{Pred}, GT answers are labeled by \underline{GT}.}
  \label{fig:wrong_examples}
\end{figure*}

\subsection{More Prediction Examples}
We show 6 correct prediction examples from STAGE in Fig.~\ref{fig:correct_examples}. As can be seen from the figure, correct examples usually have correct temporal and spatial localization. In Fig.~\ref{fig:wrong_examples} we show 6 incorrect examples. Incorrect object localization is one of the most frequent failure reason, while the model is able to localize common objects, it is difficult for it to localize unusual objects (Fig.~\ref{fig:wrong_examples}(a, d)), small objects (Fig.~\ref{fig:wrong_examples}(b)). Incorrect temporal localization is another most frequent failure reason, e.g., Fig.~\ref{fig:wrong_examples}(c, f). There are also cases where the objects being referred are not present in the sampled frame, as in Fig.~\ref{fig:wrong_examples}(e).

\end{document}